\pdfoutput=1

\documentclass[11pt]{article}

\usepackage[]{acl}

\usepackage{times}
\usepackage{latexsym}

\usepackage[T1]{fontenc}
\usepackage{amsmath}
\usepackage{amsfonts}

\usepackage[utf8]{inputenc}

\usepackage{microtype}

\usepackage{inconsolata}

\usepackage{graphicx}  
\usepackage{array}
\usepackage{multirow}  
\usepackage{CJKutf8} 
\usepackage[
]{geometry}
\usepackage{float}
\usepackage{subcaption} 
\usepackage{tabularx}
\newcolumntype{Y}{>{\centering\arraybackslash}X}

\title{Enhancing Cross-lingual Sentence Embedding for Low-resource Languages with Word Alignment}

\author{Zhongtao Miao, Qiyu Wu, Kaiyan Zhao, Zilong Wu, Yoshimasa Tsuruoka \\
  The University of Tokyo, Tokyo, Japan \\
  \texttt{\{mzt, qiyuw, zhaokaiyan1006, zw2599, yoshimasa-tsuruoka\}@g.ecc.u-tokyo.ac.jp}
\\}

\newcommand{\modelname}{WACSE}

\begin{document}
\maketitle
\begin{abstract}
The field of cross-lingual sentence embeddings has recently experienced significant advancements, but research concerning low-resource languages has lagged due to the scarcity of parallel corpora. This paper shows that cross-lingual word representation in low-resource languages is notably under-aligned with that in high-resource languages in current models. To address this, we introduce a novel framework that explicitly aligns words between English and eight low-resource languages, utilizing off-the-shelf word alignment models. This framework incorporates three primary training objectives: aligned word prediction and word translation ranking, along with the widely used translation ranking. We evaluate our approach through experiments on the bitext retrieval task, which demonstrate substantial improvements on sentence embeddings in low-resource languages. In addition, the competitive performance of the proposed model across a broader range of tasks in high-resource languages underscores its practicality.

\end{abstract}

\section{Introduction}
Cross-lingual sentence embedding encodes multilingual texts into a shared semantic embedding space in which the texts are understandable across different languages.
Various applications including bitext retrieval \citep{10.1162/tacl_a_00288} and cross-lingual semantic textual similarity tasks \citep{cer-etal-2017-semeval, chen-etal-2022-semeval} rely on cross-lingual sentence embedding.

\begin{figure}[t]
    \centering
    \includegraphics[width=\columnwidth]{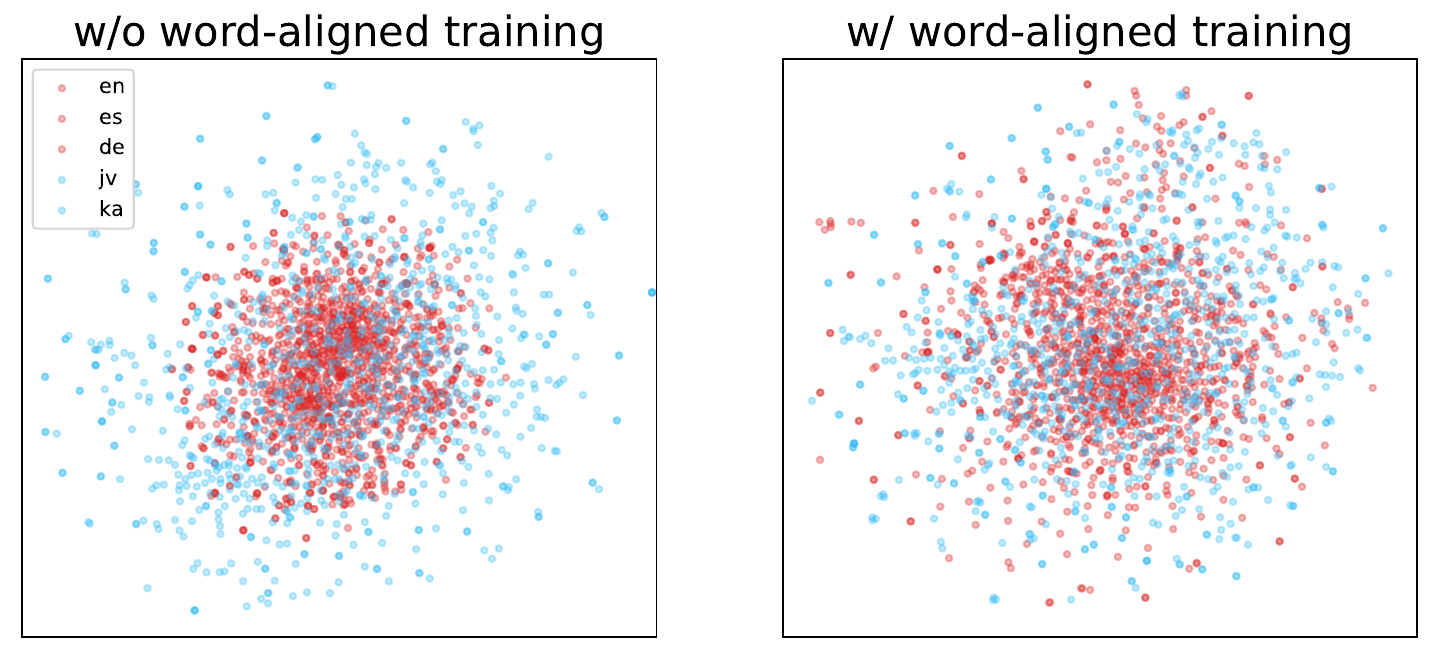}
    \caption{t-SNE visualization of sampled word embeddings from both high-resource and low-resource languages. The red points represent the word embeddings from high-resource languages, and the blue points correspond to those from low-resource languages. This comparison highlights the differences of word representation in the models w/ and w/o the explict word-aligned training.
    \textbf{Left}: words in low-resource languages are under-aligned with their translations in high-resource languages.
    \textbf{Right}: the phenomenon of under-alignment is mitigated through the proposed explicit word-aligned training.
    The details of word sampling, word embeddings and word-aligned training are described in Section~\ref{sec:impdetail}.}
    \label{fig:word_embedding_comparison}
\end{figure}

Current approaches to obtaining cross-lingual sentence embeddings primarily utilize multilingual pre-trained language models~\citep{devlin-etal-2019-bert, NEURIPS2019_c04c19c2, conneau-etal-2020-unsupervised} that employ masked language modeling and translation language modeling objectives to predict masked tokens within the context. Such models implicitly align the contextual representations of semantically similar units of sentences in different languages~\citep{li-etal-2021-multi}, thereby enabling the models to understand texts in various languages.

While the field of cross-lingual sentence embedding has recently seen great advancements~\cite{li-etal-2023-dual,li-etal-2021-multi,zhang2023veco,feng-etal-2022-language}, research concerning low-resource languages has lagged due to the scarcity of parallel corpora.



In Figure \ref{fig:word_embedding_comparison}, we observe that word embeddings from low-resource languages, which are derived from current cross-lingual models trained solely with a sentence-level alignment objective, are under-aligned with those from high-resource languages.
To address this under-alignment, we introduce a new framework featuring two word-level alignment objectives: aligned word prediction and word translation ranking.
These objectives are designed to align the word-level signals of parallel sentences. 
Additionally, a sentence-level alignment objective, known as translation ranking~\citep{feng-etal-2022-language}, is also used to ensure the basic sentence understanding. We name our proposed framework \modelname{}~(\textbf{W}ord \textbf{A}ligned \textbf{C}ross-lingual \textbf{S}entence \textbf{E}mbedding).
The right sub-figure in Figure~\ref{fig:word_embedding_comparison} shows the distribution of word embeddings obtained from the model trained with the proposed aligned word prediction and word translation ranking. It demonstrates that the under-alignment phenomenon can be mitigated through the explicitly word-aligned objectives.


The experiment results demonstrate that the proposed word-aligned training objectives can enhance cross-lingual sentence embedding, particularly for low-resource languages, as evidenced on the Tatoeba dataset~\citep{10.1162/tacl_a_00288}. This finding matches our observations on word representations in Figure~\ref{fig:word_embedding_comparison}.
Furthermore, our model retains competitive results across a broader range of tasks, including STS22~\citep{chen-etal-2022-semeval}, BUCC~\citep{zweigenbaum-etal-2017-overview}, and XNLI~\citep{conneau-etal-2018-xnli}, in which most languages are high-resource. This indicates the practicality and robustness of the proposed framework.

\section{Related Work}


\subsection{Cross-lingual Sentence Embedding}
Cross-lingual sentence embedding is the task of encoding sentences from various languages into a shared embedding space. Traditionally, large-scale parallel corpora have been utilized to learn cross-lingual sentence embeddings.
LASER \citep{artetxe-schwenk-2019-massively} employs a BiLSTM encoder trained on parallel sentences from 93 languages, totaling 223 million parallel sentences, to learn joint multilingual sentence representations.
LaBSE \citep{feng-etal-2022-language} learns cross-lingual sentence embeddings by integrating dual-encoder translation ranking, additive margin softmax, masked language modeling (MLM) and translation language modeling (TLM), utilizing training data consisting of 17 billion monolingual sentences and 6 billion translation pairs. Extending SimCSE~\citep{gao-etal-2021-simcse} to multilingual settings, mSimCSE~\citep{wang-etal-2022-english} demonstrates that contrastive learning applied to English data alone can yield universal cross-lingual sentence embeddings without the need for parallel data. Inspired by PCL~\citep{wu-etal-2022-pcl}, MPCL~\citep{zhao-etal-2024-leveraging} leverages multiple positives from different languages to improve cross-lingual sentence embedding.

\paragraph{Token-level auxiliary tasks.}
Recently, the importance of token-level auxiliary tasks has been recognized.
VECO2.0~\citep{zhang2023veco} employs thesauruses for token-to-token alignment, achieving notable results on the XTREME benchmark~\citep{pmlr-v119-hu20b}. DAP~\citep{li-etal-2023-dual} is designed with two primary objectives. The first objective, translation ranking (TR), aims to bring parallel sentences closer together in the embedding space. The second objective, representation translation learning (RTL), employs one-sided contextualized token representations to reconstruct their translation counterparts, aiming to capture the relationships between tokens in parallel sentences. TR as a simple but effective objective, is also utilized in our framework to ensure the basic sentence understanding.
Nevertheless, researchers recognize the significance of token-level or word-level alignment in cross-lingual scenarios, the acquisition of token-level or word-level supervisory signals remains a challenging topic of ongoing discussion. 
\citet{li-etal-2021-multi} employ fast\_align~\citep{dyer-etal-2013-simple} to obtain word-level supervisory signals.
XLM-Align~\citep{chi-etal-2021-improving} leverages self-labeled word alignment signals for model training.
VECO2.0~\citep{zhang2023veco} utilizes thesauruses to acquire token-level supervisory signals. 


\subsection{Word Alignment}
Word alignment is a task aimed at aligning the corresponding words in parallel sentences~\citep{brown-etal-1993-mathematics,och-ney-2003-systematic, dyer-etal-2013-simple, dou-neubig-2021-word, wu-etal-2023-wspalign}, serving as a useful component for applications such as machine translation~\citep{li-etal-2019-word, li-etal-2022-structural}.
SimAlign~\citep{jalili-sabet-etal-2020-simalign} utilizes multilingual word embeddings for word alignment without relying on parallel data or dictionaries. \citet{nagata-etal-2020-supervised}~redefine the word alignment task as a cross-lingual span prediction problem and fine-tune mBERT with manually annotated word alignment data. WSPAlign~\citep{wu-etal-2023-wspalign} reduces the dependence on manually annotated data by creating a large-scale, weakly-supervised dataset for word alignment. By pre-training word aligners with weakly-supervised signals via span prediction, it achieves state-of-the-art performance across five word alignment datasets. In this work, we employ WSPAlign to obtain the word-level supervisory signals for training models.



\section{Method}
\begin{figure*}[h]
    \centering
    \includegraphics[width=\textwidth]{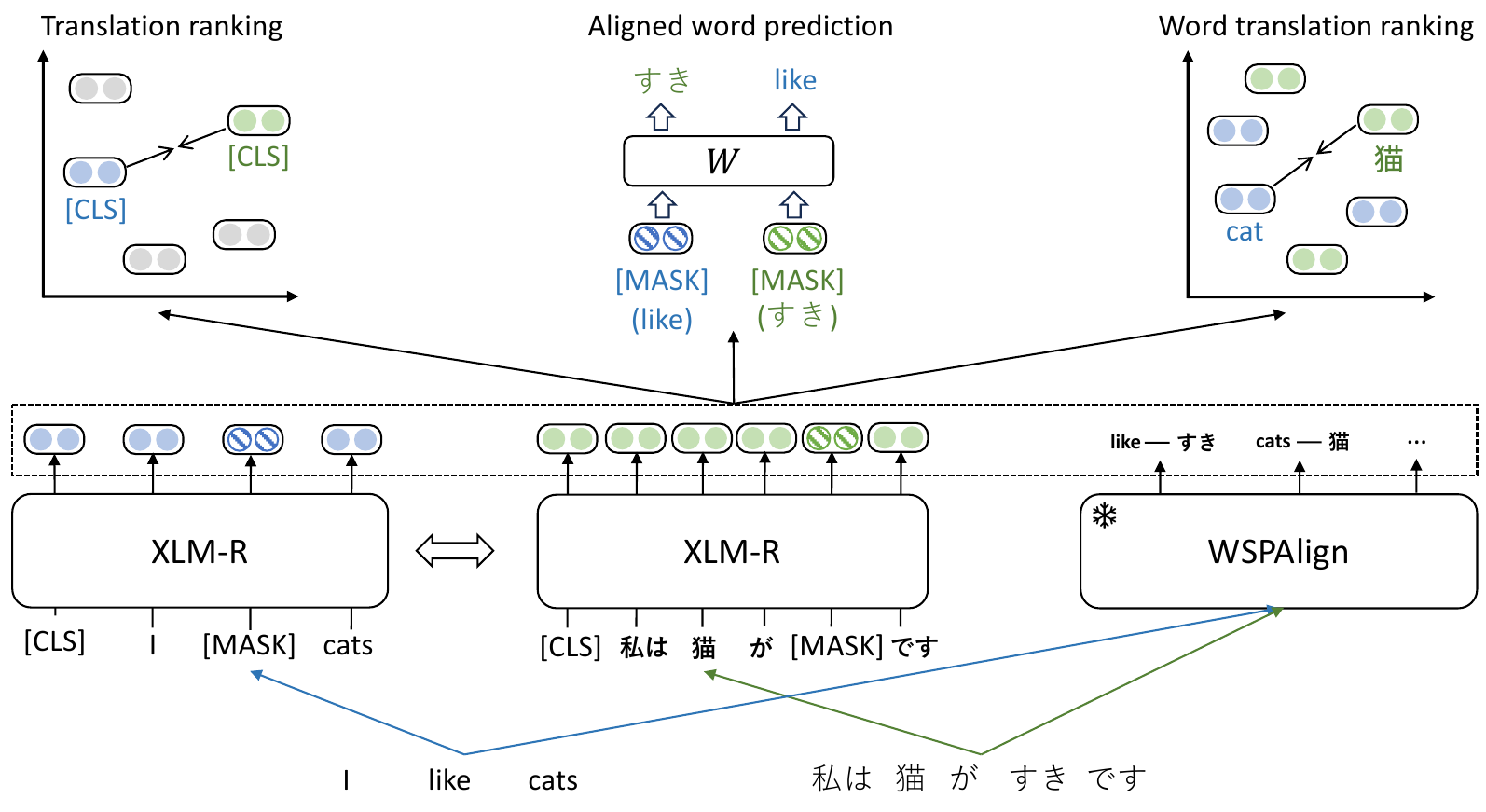}
    \caption{Illustration of \modelname{} framework. A parallel sentence pair is fed into the multilingual model along with a frozen word alignment model to obtain sentence representations, contextual token representations, and word alignment respectively. Then three objectives are calculated: (1) translation ranking: aligning sentence-level semantics; (2) aligned word prediction: utilizing the contextual representations of masked words to predict their aligned counterparts in another language; and (3) word translation ranking: aligning word-level semantics.}
    \label{fig:framework}
\end{figure*}

To enhance cross-lingual sentence embeddings of low-resource languages through explicit alignment of words, \modelname{} incorporates three tasks, translation ranking (TR), aligned word prediction (AWP) and word translation ranking (WTR) tasks. These tasks collectively aim to learn the cross-lingual sentence representations of parallel sentences. The framework is depicted in Figure~\ref{fig:framework}.

Formally, we start with a parallel dataset $(\mathbb{X}, \mathbb{Y})$ in two languages and the $i$-th parallel sentence pair is denoted as $(X_i, Y_i)$. $X_i$ and $Y_i$ can be represented as a sequence of words: $X_i = x_1, x_2, \dotsc, x_{|X_i|}$ and $Y_i = y_1, y_2, \dotsc, y_{|Y_i|}$, respectively where $|\cdot|$ denotes the length of the given sentence. After inputting a sentence into the model, we obtain the hidden representations from the last layer as follows: 
\begin{equation}
    h_{cls}^{X_i}, h_1^{X_i}, h_2^{X_i}, \dotsc, h_{|X_i|}^{X_i} = f(X_i),
\end{equation}
where $f$ represents the encoder, and $h_i^{X_i}$ denotes the corresponding hidden representation of $x_i$ in sentence $X_i$. 

Note that $h_i^{X_i}$ could be a sequence of embeddings because a word could be tokenized into multiple tokens. This could affect some minor implementation in the practice. Refer to Section~\ref{sec:impdetail} for the detailed implementation regarding this issue. Particularly, $h_{cls}^{X_i}$ is the hidden state of the \texttt{cls} token for representing the whole sentence.

\paragraph{Acquisition of Word Alignment Supervision.}
Word alignment models enable us to identify semantically equivalent word-level units within parallel sentences.
We utilize WSPAlign\footnote{\url{https://github.com/qiyuw/WSPAlign.InferEval}} to obtain the word-level supervisory signals which will be used in the calculation of AWP and WTR losses.

For the $i$-th parallel sentence $(X_i, Y_i)$, a word alignment model can generate bidirectional word pair dictionary $\textnormal{WA}^{X_i\rightarrow Y_i}$ and $\textnormal{WA}^{Y_i\rightarrow X_i}$ as follows:
\begin{equation}
\label{eqn:word_pair1}
    \textnormal{WA}^{X_i\rightarrow Y_i}, \textnormal{WA}^{Y_i\rightarrow X_i} = \textnormal{WordAlign}(X_i, Y_i).
\end{equation}


Using $\textnormal{WA}^{X_i\rightarrow Y_i}$, we can look up an aligned word $y_k \in Y_i$ for a specific $x_j \in X_i$, if it exists, and vice versa. The bidirectional dictionaries record all obtainable word pairs, demonstrated by the following equation:

\begin{equation}
\label{eqn:word_pair2}
     y_k = \textnormal{WA}^{X_i\rightarrow Y_i}(x_j), 1 \le j \le |X_i|.
\end{equation}

Here, each word pair $(x_j, y_k)$ represents a semantically equivalent word pair from the two sentences. In practice, we exclude word pairs with alignment scores below a specified threshold.
The threshold value\footnote{\url{https://github.com/qiyuw/WSPAlign.InferEval/blob/49ac6fb87fab17079153bcce84c3ac52d4ce6752/inference.py\#L74C5-L74C24}} for WSPAlign which we use is set to $0.9$.

\subsection{Aligned Word Prediction (AWP) Task}
After obtaining word alignment supervisory signals, 
we introduce AWP objective to align semantically equivalent words across different languages.  

For a word pair $(x_j, y_k)$ derived from $(X_i, Y_i)$, as introduced in Equations \ref{eqn:word_pair1} and \ref{eqn:word_pair2}, the model is tasked with predicting $y_k$ while $x_j$ is masked.

We define the aligned word prediction loss for $X_i$ as follows:

\begin{equation}
\label{eqn:mlm}
\begin{split}
    l^{AWP}(X_i) &= \sum_{x_j \in \textnormal{WA}^{X_i\rightarrow Y_i}}{\textnormal{MLM}}(X_i, x_j; y_k), \\
    &y_k = \textnormal{WA}^{X_i\rightarrow Y_i}(x_j),
\end{split}
\end{equation}
where masking language modeling ($\textnormal{MLM}$) means that the model predicts $y_k$ while masking $x_j$.
The total loss of a batch $ \mathcal{L}^{AWP}$ is given by:


\begin{equation}
    \mathcal{L}^{AWP} = \frac{1}{2N} \sum_{(X_i, Y_i) \in (\mathbb{X}, \mathbb{Y})}(l^{AWP}(X_i)+l^{AWP}(Y_i)),
\end{equation}
where $N$ is the batch size. This calculation incorporates both $X_i\rightarrow Y_i$ and $Y_i\rightarrow X_i$ directions.


\subsection{Word Translation Ranking (WTR) Task}
Besides the AWP task, previous studies have shown that token-level contrastive learning is also effective in cross-lingual pre-training~\citep{li-etal-2021-multi,zhang2023veco}.  
Inspired by this, we introduce WTR task in this section.
WTR differs from the approach taken by VECO2.0~\citep{zhang2023veco}, which utilizes thesauruses for token-to-token contrastive learning. The thesaurus-based method overlooks the contextual information of parallel sentences. In contrast, our approach leverages word pair supervision from word alignment which considers the contextual information of words in parallel sentences to align semantically equivalent tokens within parallel sentences.

For a given sentence pair $(X_i, Y_i)$ and a specific word pair $(x_j, y_k)$ obtained from it, the 
word-level WTR loss $l^{WTR}(x_j)$ for $x_j$ can be calculated as follows:

\begin{equation}
    - \log{ \frac{e^{\phi^m({h_j^{X_i}, h_k^{Y_i})})}}{e^{\phi^m{(h_j^{X_i}, h_k^{Y_i})}} + \sum_{n=1 \wedge n \neq k}^{|Y_i|} e^{\phi^m{(h_j^{X_i}, h_n^{Y_i})}}} },
\end{equation}
where $\phi^m$ particularly denotes a pair-wise cosine similarity function as the length of $h_j^{X_i}$ may not be equal to that of $h_k^{Y_i}$.
Given that the word alignment model produces multiple word pairs, the loss for the whole sentence $X_i$ is calculated as:
\begin{equation}
    l^{WTR}(X_i) = \sum_{x_j \in \textnormal{WA}^{X_i\rightarrow Y_i}}l^{WTR}(x_j).
\end{equation}

Considering bidirectional prediction across the entire batch, the loss $\mathcal{L}^{WTR}$ is presented as follows:
\begin{equation}
    \frac{1}{2N} \sum_{(X_i, Y_i) \in (\mathbb{X}, \mathbb{Y})}(l^{WTR}(X_i)+l^{WTR}(Y_i)).
\end{equation}




\subsection{Translation Ranking (TR) Task}
The dual-encoder architecture, combined with the TR task, has been shown to be effective in learning cross-lingual sentence embeddings at the sentence level, as evidenced by various studies (\citealp{guo-etal-2018-effective}; \citealp{ijcai2019p746}; \citealp{feng-etal-2022-language}).
The TR task aligns the sentence representations of different languages at the sentence level to ensure the basic sentence understanding.

Following \citet{feng-etal-2022-language} and \citet{li-etal-2023-dual}, we denote the loss of the TR task for a parallel sentence $(X_i, Y_i)$ as follows:
\begin{equation}
    l^{TR}_{i} = - \log{ \frac{e^{\phi({h_{cls}^{X_i}, h_{cls}^{Y_i})})}}{e^{\phi{(h_{cls}^{X_i}, h_{cls}^{Y_i})}} + \sum_{j=1 \wedge j \neq i}^{N} e^{\phi{(h_{cls}^{X_i}, h_{cls}^{Y_j})}}} }.
\end{equation}

For the entire batch, the total loss of TR is:
\begin{equation}
    \mathcal{L}^{TR} =  \frac{1}{N} \sum_{i=1}^{N} l^{TR}_{i},
\end{equation}
where $N$ represents the batch size, and $\phi$ denotes the cosine similarity function.
The \texttt{cls} representations, $h_{cls}^{X_i}$ and $h_{cls}^{Y_i}$, are used to calculate the similarity between $X_i$ and $Y_i$. 

The final loss is calculated as the weighted sum of three losses:
\begin{equation}
    \mathcal{L} =  \alpha\mathcal{L}^{TR} + \beta\mathcal{L}^{AWP} + \gamma\mathcal{L}^{WTR},
\end{equation}
where $\alpha$ is the weight for the TR loss, $\beta$ for the AWP loss, and $\gamma$ for the WTR loss. 


\section{Experimental Setup}
\subsection{Training Data}

We utilize the same parallel corpora for training as DAP~\citep{li-etal-2023-dual}, which is English-centric and  comprises 36 language pairs.
We use ISO 639 language codes\footnote{\url{https://en.wikipedia.org/wiki/List_of_ISO_639_language_codes}} (two-letter codes) to denote languages.
Using the same dataset as DAP, we employ WSPAlign to identify word-level semantically equivalent units.
The statistics of the parallel corpora we use are presented in Table \ref{table:training_data}.

\begin{table}[h]
\centering
\small
\begin{tabularx}{\columnwidth}{Y|Y|Y}
\noalign{\hrule height 0.8pt}
\hline
\multirow{2}{*}{Language Pair} &  \multicolumn{2}{c}{\# Parallel Sentences} \\
\cline{2-3}
 & train & dev \\
\noalign{\hrule height 0.6pt}
\hline
en-kk & 18190 & 2021\\
\hline
en-te & 78105 & 8678 \\
\hline
en-ka & 146905 & 10K\\
\hline
en-jv & 317252 & 10K \\
\hline
en-other & 1M & 10K \\
\noalign{\hrule height 0.8pt}
\hline
\end{tabularx}
\caption{Number of parallel sentences per language in the training and development corpora.}
\label{table:training_data}
\end{table}

\begin{table}[h]
\centering
\resizebox{\columnwidth}{!}{%
\begin{tabular}{c|c|c|c|c}
\noalign{\hrule height 0.8pt}
\hline
Lang. Code &  \# Articles & & Lang. Code & \# Articles\\ 
\cline{1-2} \cline{4-5}
tl & 45750 & & jv & 72851 \\ \cline{1-2} \cline{4-5}
sw & 78915 & & ml & 84939 \\ \cline{1-2} \cline{4-5}
te & 88914 & & mr & 94005 \\ \cline{1-2} \cline{4-5}
af & 113208 & & bn & 144218 \\ \cline{1-2} \cline{4-5}
hi & 159888 & & th & 160499\\ \cline{1-2} \cline{4-5}
ta & 160712 & & ka & 169878 \\ \cline{1-2} \cline{4-5}
ur & 198346 & & el & 228223\\ \cline{1-2} \cline{4-5}
kk & 235611 & & et & 241085 \\ \cline{1-2} \cline{4-5}
bg & 294740 & & he & 345544 \\ \cline{1-2} \cline{4-5}
eu & 424058 & & hu & 533933 \\ \cline{1-2} \cline{4-5}
tr & 540433 & & fi & 563464 \\ \cline{1-2} \cline{4-5}
ko & 652657 & & id & 673857\\ \cline{1-2} \cline{4-5}
fa & 983682 & & pt & 1114362\\ \cline{1-2} \cline{4-5}
ar & 1223016 & & vi & 1289408\\ \cline{1-2} \cline{4-5}
zh & 1390659 & & ja & 1395361\\ \cline{1-2} \cline{4-5}
it & 1838179 & & es & 1911915\\ \cline{1-2} \cline{4-5}
ru & 1950729 & & nl & 2141291\\ \cline{1-2} \cline{4-5}
fr & 2573743 & & de & 2859124\\ \cline{1-2} \cline{4-5}
\noalign{\hrule height 0.8pt}
\hline
\end{tabular}
}
\caption{Number of Wikipedia articles available per language in the 36 languages of the training parallel corpora (accessed time: 2023-12-05 17:21:49).}
\label{table:wikepedia_articles}
\end{table}

\begin{table*}
    \centering
    \begin{tabular}{c|cccc}
        \noalign{\hrule height 0.8pt}
        \hline
        Model & 4 langs & 5 langs & 8 langs & 36 langs \\
        \hline
        LaBSE & 92.5 & 93.5 & 93.8 & 95.4 \\
        InfoXLM & 35.4 & 32.8 &  39.3 & 57.0 \\
        
        \hline
         DAP & 73.9 & 73.4 & 79.6 & 92.0 \\
         \modelname{}~\textbf{(ours)} & \textbf{75.9\small{(+2.0)}} & \textbf{76.0\small(+2.6)} & \textbf{81.2\small{(+1.6)}} & \textbf{92.1\small{(+0.1)}}  \\
        \noalign{\hrule height 0.8pt}
        \hline
    \end{tabular}
    \caption{Average accuracy on the Tatoeba dataset across both directions for selected languages. The chosen low-resource languages are (kk, te, ka, jv) for ``4 langs'', (tl, jv, ka, kk, te) for ``5 langs'', and (te, ka, kk, jv, ml, sw, tl, mr) for ``8 langs''. Results of DAP are from~\citet{li-etal-2023-dual}.}
    \label{table:tatoeba-num-langs}
\end{table*}


\subsection{Low-resource Languages\label{sec:low-resource-lang}}
We focus on cross-lingual sentence embeddings in low-resource languages. 
In the experiment, we determine low-resource languages based on two criteria: (1) the number of Wikipedia articles available per language\footnote{\url{https://meta.wikimedia.org/wiki/List_of_Wikipedias}} and (2) the size of the training data available for each language.
Among the 36 languages in our dataset, six languages (tl, jv, sw, ml, te, mr) are identified as low-resource based on the smallest number of Wikipedia articles, according to criterion 1.
For criterion 2, we select four languages (kk, te, ka, jv) with the fewest parallel sentences in the training set.  
Detailed information on the number of Wikipedia articles per language is available in Table~\ref{table:wikepedia_articles}.
Considering the intersection of two criteria, we classify eight languages as low-resource in this study. Furthermore, we assess our proposed approach using various combinations of these eight languages, including settings with four languages (kk, te, ka, jv), five languages (tl, kk, te, ka, jv) and all eight languages.


\subsection{Implementation Details}
\label{sec:impdetail}
As we mentioned above, $h_j^{X_j}$ and $h_k^{Y_k}$ could consist of multiple hidden states and the number of them could be different. When calculating MLM loss in Equation~\ref{eqn:mlm}, we roughly clip the longer sequence of tokens to ensure the number of tokens are equivalent.

As for the details of Figure~\ref{fig:word_embedding_comparison}, the words are sampled from the training dataset based on their frequencies, totaling 500 words. These word embeddings are extracted from the embedding layers of XLM-R models. The left sub-figure illustrates the result from the model trained solely with the TR objective, while the right sub-figure displays the result of our model, which is trained using the proposed method.

\paragraph{Model Size.}
For the Transformer encoder model~\citep{NIPS2017_3f5ee243} that we use, we adopt the configuration of XLM-R model~\citep{conneau-etal-2020-unsupervised}.
We initialize the encoder model using the \texttt{xlm-roberta-base} checkpoint\footnote{\url{https://huggingface.co/xlm-roberta-base}}.

\paragraph{Hyperparameters.}
The maximum sequence length is set to 32.
We train our model using the AdamW optimizer, with a learning rate of 5e-5. The training steps are 10K or 100K depending on different evaluation tasks.
Gradient accumulation is employed across two A100 GPUs, resulting in a total batch size of 1024. The reported results are the average of two random seeds (42 and 0). The values of $\alpha$, $\beta$ and $\gamma$ are set to 0.8, 0.1 and 0.1 empirically.
For all models, the pooling method is configured as \texttt{cls\_before\_pooler}.


In line with DAP, we evaluate the model every 2,000 steps using development set shown in Table~\ref{table:training_data}. Similarity search, which is a widely-used metric in cross-lingual retrieval tasks~\citep{10.1162/tacl_a_00288}, is utilized for choosing the optimal checkpoint.

\subsection{Baselines}
We compare our proposed method with XLM-R and its TR fine-tuned variant. Other competitive models such as InfoXLM~\citep{chi-etal-2021-infoxlm}, LaBSE~\citep{feng-etal-2022-language}, and mSimCSE~\citep{wang-etal-2022-english} are also included in the comparison.
Note that some of these models leverage significantly larger datasets than ours. For instance, LaBSE utilizes 17 billion monolingual sentences and 6 billion translation pairs, while ours is only in the scale of 36 million. 

Our main baseline is DAP~\citep{li-etal-2023-dual}, which is a recent cross-lingual sentence embedding model leveraging token-level information.
Hence, We adopt the identical settings including training data, model size, and other hyperparameters\footnote{\url{https://github.com/ChillingDream/DAP}}. 

\section{Evaluation Tasks and Results}
\subsection{Bitext Retrieval}
Bitext retrieval is the task of retrieving the most relevant sentence from a target language corpus given a query sentence in the source language~\citep{li-etal-2023-dual}. The Tatoeba dataset~\citep{10.1162/tacl_a_00288} is a benchmark for evaluating bitext retrieval spanning a broad array of languages. We train our model for 100K steps and evaluate it on Tatoeba in this task.
The released checkpoints of LaBSE\footnote{\url{https://huggingface.co/sentence-transformers/LaBSE}} and InfoXLM\footnote{\url{https://huggingface.co/microsoft/infoxlm-base}} are used for comparison.

\paragraph{Results.} We report the results of the Tatoeba dataset across four settings, as detailed in Table~\ref{table:tatoeba-num-langs}. The low-resource language settings, including the four-language, five-language, and eight-language settings, are described in Section~\ref{sec:low-resource-lang}. The thirty-six-language setting encompasses all 36 languages in the training dataset. From Table~\ref{table:tatoeba-num-langs}, we can observe that our model improves the cross-lingual sentence embedding in all low-resource language settings. But when expanding to all 36 languages, the improvement becomes marginal.

A possible explanation for this is that current cross-lingual sentence embedding models may struggle with learning the word-level alignment in low-resource languages due to the limited training data available. Through the explicit word-level alignment objectives, our method facilitates the alignment of the semantically equivalent tokens between high-resource languages and low-resource languages, aiding the model in acquiring basic word-level semantic information for low-resource languages. Therefore, The proposed method can improve cross-lingual sentence embeddings of low-resource languages. In contrast, high-resource languages already achieve effective word-level alignment during the pre-training phase with implicit word-level signals in the rich parallel corpus.
Hence, continuing to explicitly align word-level semantic units between two high-resource languages could detract from the language-dependent and sentence-level features of the cross-lingual sentence embeddings.

\subsection{Cross-lingual Semantic Textual Similarity}
Semantic Textual Similarity (STS) assesses the degree of similarity between two sentences. The cross-lingual STS task expands this to multilingual scenarios. For this task, we utilize STS22~\citep{chen-etal-2022-semeval} dataset and evaluate the performance using the MTEB benchmark (version 1.1.1)~\citep{muennighoff-etal-2023-mteb}. According to MTEB, the Spearman correlation, based on similarity, is the chosen metric for evaluation~\citep{reimers-etal-2016-task}. 
For both DAP and our method, we train the model for 10K steps and test on STS22. \citet{zhao-etal-2024-leveraging} point out that the result for $fr \leftrightarrow pl$ (French-Polish) language pair in STS22 seems unstable. Consequently, we report two versions of the STS22 task results, one including all language pairs in the STS22 dataset of MTEB benchmark and the other excluding the $fr \leftrightarrow pl$ pair.

\paragraph{Results.}  As shown in Table~\ref{table:avg_sts22}, our method significantly outperforms DAP on the STS22 dataset. This improvement illustrates that leveraging word-level semantically equivalent units, obtained through word alignment, can enhance the performance of cross-lingual sentence embedding models on cross-lingual STS tasks. This enhancement occurs by bringing semantically equivalent units closer across languages, even though the languages in STS22 are not considered low-resource.
It is noteworthy that LaBSE performs slightly better than \modelname{} on STS22.
Though LaBSE utilizes a much larger training dataset, \modelname{} still achieves competitive results.
Detailed scores of STS22 are provided in Table~\ref{table:sts22}.

\begin{table}[h]
    \centering
    \small
    \begin{tabularx}{\columnwidth}{c|YY}
        \noalign{\hrule height 0.8pt}
        \hline
        \multirow{2}{*}{Model} &   \multicolumn{2}{c}{STS22} \\
        \cline{2-3}
         &  Avg. & Avg.(-fr-pl) \\
        \hline
        LaBSE &  59.2 & 59.1 \\
        InfoXLM & 49.6 & 47.5 \\
        \hline
        XLM-R+DAP &  51.7 & 52.1  \\
         XLM-R+ \modelname{} & \textbf{58.7\small{(+7.0)}} & \textbf{58.5\small{(+6.4)}} \\
        \noalign{\hrule height 0.8pt}
        \hline
    \end{tabularx}
    \caption{Spearman correlation scores of STS22. The results of LaBSE and InfoXLM are obtained using the MTEB benchmark.}
    \label{table:avg_sts22}
\end{table}

\begin{table}[h]
    \centering
    \resizebox{1\columnwidth}{!}{%
    \begin{tabular}{c|cccccc}
    \noalign{\hrule height 0.8pt}
    \hline
             & ar   & de    & de-en & de-fr & de-pl & en    \\ \hline
XLM-R + DAP  & 49.2 & 38.0  & 43.3  & 49.8  & 43.6  & 55.2  \\
XLM-R + Ours & 55.2 & 41.6  & 47.9  & 52.2  & 49.9  & 60.2  \\ \hline
             & es   & es-en & es-it & fr    & fr-pl & it    \\ \hline
XLM-R + DAP  & 59.0 & 62.1  & 55.1  & 67.1  & 45.0 & 66.2  \\
XLM-R + Ours & 60.0 & 70.0  & 65.5  & 73.4  & 62.0  & 71.2  \\ \hline
             & pl   & pl-en & ru    & tr    & zh    & zh-en \\ \hline
XLM-R + DAP  & 30.0 & 55.1  & 49.8  & 50.0  & 58.3  & 54.6  \\
XLM-R + Ours & 33.8 & 69.1  & 55.0  & 57.7  & 63.2  & 68.9  \\ \noalign{\hrule height 0.8pt} \hline
    \end{tabular}%
    }
    \caption{Detailed results of the STS22 dataset.}
    \label{table:sts22}
\end{table}

\subsection{Bitext Mining}
Bitext mining involves extracting parallel sentences from two monolingual corpora with the assumption that some of these sentences are translation pairs. Following the settings of DAP and mSimCSE, we assess our model using the BUCC dataset~\citep{zweigenbaum-etal-2017-overview} which includes four language pairs: $fr \leftrightarrow en$, $de \leftrightarrow en$, $ru \leftrightarrow en$ and $zh \leftrightarrow en$. We train our model for 10K steps and use the evaluation code from mSimCSE\footnote{\url{https://github.com/yaushian/mSimCSE}}.

\paragraph{Results.} Table~\ref{table:bucc} shows the results of different models which we compare. The results of LASER~\citep{10.1162/tacl_a_00288}, mSimCSE, XLM-R, and LaBSE are from the mSimCSE paper~\citep{wang-etal-2022-english}.
The notation ``mSimCSE$_{sw,fr}$+NLI'' refers to the variant of mSimCSE trained with a combination of English Natural Language Inference (NLI) data and translation pairs in English-Swahili and English-French~\citep{wang-etal-2022-english}.
Our proposed method outperforms the ``mSimCSE$_{sw,fr}$+NLI'' model, even it is a large size model.
From Table~\ref{table:bucc}, we can see that our approach achieves competitive results, positioning it between the performance of ``mBERT + DAP'' and ``XLM-R + DAP'' at the base model size. 
\begin{table*}[t]
    \centering
    \setlength\tabcolsep{2pt}
    \resizebox{\textwidth}{!}{
        \begin{tabularx}{\textwidth}{c|YYY|YYY|YYY|YYY|c}
        \noalign{\hrule height 0.8pt}
        \hline
        \multirow{2}{*}{Model}  & \multicolumn{3}{c}{fr-en} & \multicolumn{3}{c}{de-en} & \multicolumn{3}{c}{ru-en} & \multicolumn{3}{c}{zh-en} & avg. \\
        \cline{2-14}
             & P & R & F & P & R & F & P & R & F & P & R & F & F \\
        \hline
        LASER & $-$ & $-$ & $-$ & $-$ & $-$ & $-$ & $-$ & $-$ & $-$ & $-$ & $-$ & $-$ & 92.9 \\
        \hline
        \multicolumn{14}{c}{XLM-R large} \\
        \hline
        mSimCSE$_{sw,fr}$ + NLI  & $-$ & $-$ & $-$ & $-$ & $-$ & $-$ & $-$ & $-$ & $-$ & $-$ & $-$ & $-$ &  93.6 \\
        \hline

        \multicolumn{14}{c}{XLM-R base} \\
        \hline
        XLM-R & $-$ & $-$ & $-$ & $-$ & $-$ & $-$ & $-$ & $-$ & $-$ & $-$ & $-$ & $-$ & 66.0 \\
        \hline
        
        LaBSE & $-$ & $-$ & $-$ & $-$ & $-$ & $-$ & $-$ & $-$ & $-$ & $-$ & $-$ & $-$ & 93.5 \\
        \hline
        mBERT + DAP  & 94.1 & 92.9 & 93.5 & 97.5 & 93.8 & 95.6 & 96.7 & 90.8 & 93.7 & 94.5 & 93.2 & 93.8 & 94.1 \\
        XLM-R + DAP  & 94.1 & 93.2 & 93.7 & 97.5 & 95.6 & 96.5 & 97.8 & 94.2 & 96.0 & 96.4 & 93.6 & 95.0 & \textbf{95.3} \\
        XLM-R + \modelname{} \textbf{(ours)}  & 93.8 & 93.4 & 93.6 & 97.9 & 94.9 & 96.4 & 97.0 & 94.0 & 95.4 & 94.1 & 95.3 & 94.7 & 95.0\small{(-0.3)} \\ 
        
        \noalign{\hrule height 0.8pt}
        \hline
        \end{tabularx}
    }
    \caption{Performance on the BUCC dataset. ``mBERT + DAP'' and ``XLM-R + DAP''~\citep{li-etal-2023-dual} are our re-implemented results with the same 10K training steps as ``XLM-R + \modelname{}''. The results of LaBSE are from mSimCSE paper~\citep{wang-etal-2022-english}.}
    \label{table:bucc}
\end{table*}

\begin{table*}[t]
\setlength\tabcolsep{2pt}
\centering
\resizebox{\textwidth}{!}{%
\begin{tabular}{l|cccccccccccccccc}
\noalign{\hrule height 0.8pt}
\hline
Model & en & fr & es & de & el & bg & ru & tr & ar & vi & th & zh & hi & sw & ur & avg. \\
\hline
InfoXLM & 86.4 & 80.3 & 80.9 & 79.3 & 77.8 & 79.3 & 77.6 & 75.6 & 74.2 & 77.1 & 74.6 & 77.0 & 72.2 & 67.5 & 67.3 & 76.5 \\
LaBSE & 85.4 & 80.2 & 80.5 & 78.8 & 78.6 & 80.1 & 77.5 & 75.1 & 75.0 & 76.5 & 69.0 & 75.8 & 71.9 & 71.5 & 68.1 & 76.3 \\
\hline
XLM-R & 83.8 & 77.6 & 78.2 & 75.4 & 75.0 & 77.0 & 74.8 & 72.7 & 72.0 & 74.5 & 72.1 & 72.9 & 69.6 & 64.2 & 66.0 & 73.7 \\
\ \ \ \ \  \  + TR & 83.5 & 76.4 & 76.8 & 75.7 & 74.2 & 76.2 & 74.6 & 71.8 & 71.1 & 74.2 & 69.1 & 72.9 & 68.8 &  66.8 & 65.2 & 73.1 \\
\ \ \ \ \  \ + TR + TLM & 84.6 & 77.4 & 76.9 & 74.9 & 68.1 & 69.8 & 69.4 & 68.1 & 61.7 & 68.9 & 62.6 & 66.9 & 61.4 & 61.7 & 57.5 & 68.7 \\
\ \ \ \ \  \  + DAP & 82.9 & 77.0 & 77.7 & 75.7 & 75.2 & 76.0 & 74.7 & 73.1 & 72.5 & 74.2 & 71.9 & 73.0 & 69.8 & 70.5 & 66.0 & 74.0 \\
\ \ \ \ \  \  + \modelname{}~\textbf{(ours)} & 83.8 & 77.7 & 78.2 & 76.5 & 75.4 & 77.2 & 75.0 & 73.1 & 72.1 & 74.9 & 72.3 & 73.5 & 69.9 & 69.0 & 65.0 & \textbf{74.2\small{(+0.2)}} \\ 
\noalign{\hrule height 0.8pt}
\hline
\end{tabular}
}
\caption{Accuracy on the XNLI dataset. Results of DAP~\citep{li-etal-2023-dual}, InfoXLM~\citep{chi-etal-2021-infoxlm} and LaBSE~\citep{feng-etal-2022-language} are taken from DAP paper~\citep{li-etal-2023-dual}.}
\label{table:xnli}
\end{table*}
\subsection{Cross-lingual Natural Language Inference}
The Cross-lingual Natural Language Inference (XNLI)~\citep{conneau-etal-2018-xnli}) is a task that requires the model to classify sentence pairs across 15 languages into categories of entailment, neutrality, and contradiction. Following the settings of~\citet{chi-etal-2021-infoxlm} and~\citet{li-etal-2023-dual}, we apply a cross-lingual transfer approach where the model is fine-tuned on English training data and then evaluated on test datasets in other languages. We use the same hyperparameter setting as DAP, with a batch size of 256 and a maximum sequence length of 128 tokens. The number of epochs is set to 2. We do not employ weight decay and experiment with learning rates of \{1e-5, 3e-5, 5e-5, 7e-5\}. The optimal learning rate is 7e-5 for our model.

\paragraph{Results.} Table~\ref{table:xnli} shows the accuracy results. 
The XNLI task does not inherently depend on cross-lingual sentence embedding, thus not directly benefiting from the training in a straightforward manner, but our model demonstrates a slight improvement over the DAP model.
Unlike DAP, which utilizes the Representation Translation Learning (RTL) objective to understand token-level relationships between parallel sentences, our model employs a novel framework with two word-level alignment objectives to align semantically equivalent token representations across languages.
This suggests that our framework's approach may offer marginal advantages over the RTL loss used by DAP in capturing the nuanced semantics necessary for cross-lingual natural language inference.
Note that this task does not directly pertain to cross-lingual sentence embedding. As a result, this observation also illustrate the practicality of our framework.

\section{Analysis}
In this section, we carry out experiments to gain a deeper understanding of the proposed framework, specifically investigating the role of language identification information and the three losses in \modelname. Our primary focus is on the Tatoeba dataset as it is the only one that encompasses the low-resource language setting. All models discussed in this section are trained with a fixed seed (42) and the training step is 10K.

\subsection{Does the Language Identification Information Matter?\label{sec:lang_embed}}
We conduct experiments to determine whether incorporating language-specific information can enhance cross-lingual sentence embeddings of low-resource languages.
Specifically, we add a new embedding layer that encodes language IDs as the language embedding. We assign different ID number for different languages. It is then added to the token embeddings of our models. This approach is designed to assess the significance of language identification information for our method. We initialize the language embedding layer randomly at the start of training. The final embedding fed into the models is the sum of the token embedding, the positional embedding and the language embedding. 

According to the results presented in Table~\ref{tab:lang_embed}, for low-resource languages, incorporating language identification information proves to be beneficial. However, for the cross-lingual sentence embeddings of all 36 languages, it appears more advantageous not to include the language identification information. 

\begin{table}[h]
    \centering
    \begin{tabular}{c|cc}
        \noalign{\hrule height 0.8pt}
        \hline
        Model &  5 langs & 36 langs\\
        \hline
        \modelname{} (w/o lang embed) & 77.4 & 91.1 \\
        \modelname{} (w/ lang embed) & 78.2 & 90.8 \\
        \noalign{\hrule height 0.8pt}
        \hline
    \end{tabular}
    \caption{Average accuracy across two directions on the Tatoeba dataset for five low-resource languages and all 36 languages. ``w/ lang embed'' denotes models trained with the language embedding layer, while ``w/o lang embed'' refers to models without this layer.}
    \label{tab:lang_embed}
\end{table}

\subsection{Do Word-level Objectives Matter?}
We also train models exclusively on the TR task to highlight the effectiveness of the AWP and WTR objectives. Following Section~\ref{sec:lang_embed}, we present results for both the low-resource language setting and the 36-language setting. As indicated in Table~\ref{tab:only_tr}, the AWP and WTR objectives prove to be effective in both scenarios. Note that their performance in the low-resource language setting surpasses that in the 36-language setting. 

\begin{table}[h]
    \centering
    \begin{tabularx}{\columnwidth}{Y|YY}
        \noalign{\hrule height 0.8pt}
        \hline
        Model &  5 langs & 36 langs\\
        \hline
        TR & 75.8 & 90.7 \\
        \modelname{} & 77.4 & 91.1 \\
        \noalign{\hrule height 0.8pt}
        \hline
    \end{tabularx}
    \caption{Average accuracy for the both directions on the Tatoeba dataset across five low-resource languages and all 36 languages. ``TR'' represents the model trained solely with the translation ranking objective, while \modelname{} refers to the model trained with the TR, AWP and WTR objectives.}
    \label{tab:only_tr}
\end{table}

\subsection{Can AWP and WTR be Used Solely?}
We present the results of models trained with the TR and AWP objectives, the TR and WTR objectives and a combination of the three objectives (TR, AWP and WTR). To accurately investigate the effect of AWP and WTR, we conduct grid search to find the optimal hyperparameters for the model trained with the combined three objectives. Specifically, the loss weights are 0.8, 0.02 and 0.18, for TR, AWP and WTR in the WACSE in Table~\ref{tab:three_objectives}, respectively.

As illustrated in Table~\ref{tab:three_objectives}, both AWP and WTR contributes to enhancing the cross-lingual sentence embeddings for low-resource languages in comparison to the model utilizing only the TR objective. Moreover, WTR exhibits a marginally superior capability for learning cross-lingual sentence embeddings than AWP. The advantage may stem from WTR's strategy of aligning word-level equivalent units within the context of parallel sentences, whereas AWP focuses on predicting masked tokens using the context of monolingual sentences. The optimal result is the combination of three objectives, showing the effectiveness of our \modelname{} framework.
\begin{table}[h]
    \centering
    \begin{tabularx}{\columnwidth}{Y|Y}
        \noalign{\hrule height 0.8pt}
        \hline
        Model & 8 langs \\
        \hline
        TR &  79.8 \\
        TR + AWP & 80.8\\
        TR + WTR & 81.1\\
        \modelname{} & 81.2\\
        \noalign{\hrule height 0.8pt}
        \hline
    \end{tabularx}
    \caption{Average accuracy across both directions on the Tatoeba benchmark dataset for the eight low-resource language setting. ``TR'' indicates the model trained exclusively with the translation ranking (TR) objective. ``TR + AWP'' refers to the model trained with both the TR and AWP objectives. ``TR + WTR'' represents the model trained with the TR objective and the WTR objective. \modelname{} denotes the model trained with a combination of the TR, AWP and WTR objectives.}
    \label{tab:three_objectives}
\end{table}

\section{Conclusion}
In this paper, we observe an intriguing phenomenon: the distributions of word embeddings of low-resource languages are under-aligned with those of high-resource languages in current multilingual pre-trained language models.
Based on this observation, we propose a framework designed to align word-level semantically equivalent units in parallel sentences between high-resource languages and low-resource languages, thereby enhancing the cross-lingual sentence embeddings for low-resource languages. Furthermore, we demonstrate that aligning word-level semantically units between two high-resource languages with our proposed method may  detrimentally affect the language-specific features learned during the pre-training phase.
Our experimental results show the effectiveness of our method in improving cross-lingual sentence embeddings for low-resource languages. Additionally, \modelname{} preserves the performance of the model on other tasks that involve high-resource languages.

\newpage

\section*{Limitations}
Our approach does not consider phrase-level alignment between high-resource languages and low-resource languages, an aspect that merits further investigation.
The effectiveness of our proposed method is significantly influenced by the quality of the word alignment model, i.e., WSPAlign. The released WSPAlign was not trained for low-resource languages particularly. Thus, developing a word alignment model with strong cross-lingual transferability is an important future direction.

\section*{Ethics Statement}

All datasets and checkpoints used in this paper are copyright free for research purpose. Previous studies are properly cited and discussed. This research aims to improve cross-lingual sentence embedding models for low-resource languages. We do not introduce additional bias to particular communities. We utilized LLM only for proofreading but not generating any specific contents in this paper.


\bibliography{anthology,custom}

\appendix

\end{document}